\pdfoutput=1

\documentclass[11pt]{article}

\usepackage{acl}

\usepackage{times}
\usepackage{latexsym}

\usepackage[T1]{fontenc}

\usepackage[utf8]{inputenc}

\usepackage{microtype}

\usepackage{tikz}
\usepackage{cite}
\usepackage{amsmath,amssymb,amsfonts}
\usepackage{graphicx}
\usepackage{textcomp}
\usepackage{xcolor}
\usepackage{mathtools}
\usepackage{enumitem}
\usepackage{bbm}
\usepackage{amsthm}
\usepackage{adjustbox}
\usepackage{multirow}
\DeclareMathOperator*{\argmax}{arg\,max}

\microtypecontext{spacing=nonfrench}

\newcommand{\red}[1]{{\color{black}#1}}

\title{Privacy Leakage in Text Classification: \\ A Data Extraction Approach}

\author{Adel Elmahdy\thanks{\phantom{b}This work was carried out as part of an internship at Microsoft Research (MSR), Redmond, WA.} \phantom{bbbb}\\
  University of Minnesota \phantom{bbbb}\\
  \texttt{adel@umn.edu} \phantom{bbbb}\\\And
  {Huseyin A.\ Inan \and Robert Sim} \\
  Microsoft Research \\
  \texttt{huseyin.inan@microsoft.com} \\
  \texttt{rsim@microsoft.com} \\
  }

\begin{document}
\maketitle

\begin{abstract}
Recent work has demonstrated the successful extraction of training data from generative language models.
However, it is not evident whether such extraction is feasible in text classification models since the training objective is to predict the class label as opposed to next-word prediction.
This poses an interesting challenge and raises an important question regarding the privacy of training data in text classification settings.
Therefore, we study the potential privacy leakage in the text classification domain by investigating the problem of unintended memorization of training data that is not pertinent to the learning task.
We propose an algorithm to extract missing tokens of a partial text by exploiting the likelihood of the class label provided by the model.
We test the effectiveness of our algorithm by inserting canaries into the training set and
attempting to extract tokens in these canaries post-training. 
In our experiments, we demonstrate that successful extraction is possible to some extent. This can also be used as an auditing strategy to assess any potential unauthorized use of personal data without consent.
\end{abstract}

\section{Introduction}
Tremendous progress has recently been made in deep learning with natural language processing (NLP), which has led to significant advances in the model performance of a wide variety of NLP applications. 
The Transformer model \citep{vaswani2017attention,wolf2020transformers} has become the central and dominant architecture of many state-of-the-art NLP models. 
However, NLP models trained with personal data have also been shown to be vulnerable to fairness \citep{mehrabi2021survey} and privacy \citep{mireshghallah2020privacy} issues, leading to adverse societal and ethical consequences.

One of the prime challenges of training machine learning models is the phenomenon of memorizing unique or rare training data. 
This may occur via what is called \textit{unintended memorization} \citep{carlini2019secret} where the trained model memorizes out-of-distribution data in the training set that is irrelevant to the learning task.
It is known that overfitting is not the cause of such a phenomenon, since the out-of-distribution data can be memorized as long as the model is still learning, making it challenging to mitigate through methods preventing overfitting such as early stopping.
This phenomenon raises privacy concerns when the training set includes private data that may be inadvertently leaked, e.g., \citep{xkcd_pm}.

The main focus of our work is to explore the memorization of training data in text classification models, which may contain private information collected from individuals.
A motivating example in our study is a topic classification setting in which an individual can have private information, such as ``I vote for X party'' in the \textit{politics} category, which can lead to a privacy violation if this information is leaked by the model.

We propose a data extraction algorithm to recover missing tokens of a partial text using the target model.
The algorithm exploits the likelihood that the model generates for the target label of the text 
to infer the unknown tokens of the partial input text.
To the best of our knowledge, this work is the first to demonstrate privacy leakage in a text classification setting by extracting tokens of canary sequences\footnote{\red{Canary sequences are out-of-distribution examples inserted into the training data. The trained model is then assessed to measure the degree to which the model has memorized such sequences.}} via access to the underlying classification model.
We conduct experiments to evaluate the performance of our extraction algorithm under a wide range of parameters such as the number of extracted tokens, the number of canary insertions, and the number of guesses for the extraction.


\section{Background: Language Modeling}
\label{sec:language_model}
\red{While this work is about text classification setting, it is built upon language models. In this section, we give a brief overview of langauge modeling.}
Language models are one of the pillars of state-of-the-art natural language processing pipelines. It has been well established that training these models at scale on large public corpora makes them adaptable to a wide range of downstream tasks \citep{foundation_models}.

Two widely used pre-training objectives are auto-regressive (AR) language modeling \citep{RadfordNSS18, RadfordWCLAS19}, and masked language modeling (MLM) \citep{DevlinCLT19, LiuOGDJCLLZS19}. AR language modeling is based on modeling the probability distribution of a text corpus by decomposing it into conditional probabilities of each token given the previous context. Specifically, the distribution $\mathbb{P} \left(x_1, x_2, \ldots, x_n\right)$ of a sequence of tokens $\left(x_1, x_2, \ldots, x_n\right)$ 
can be factorized as $\mathbb{P} \left(x_1, x_2, \ldots, x_n\right) = \Pi_{i=1}^{n} \mathbb{P} \left(x_i | x_1, x_2, \ldots, x_{i-1}\right)$
using the Bayes rule. A neural network is then trained to model each conditional distribution. We note that such a decomposition only captures the unidirectional context.

On the other hand, the MLM pre-training objective can utilize the bidirectional context since it is based on replacing a certain portion of tokens by a special symbol \texttt{[MASK]} and the model is trained to recover the original tokens at these corrupted positions. This bidirectional context information often carries useful signal on downstream language understanding tasks such as text classification tasks, leading to improved performance for models trained with MLM pre-training objective.

\section{Related Work}
The ultimate goal of training language models is to model the underlying distribution of a language, which should not require the memorization of training samples. However, recent results have shown that such memorization occurs in language models \citep{carlini2019secret, Zanella20, carlini2021extracting, inan2021training, mireshghallah2021privacy, carlini22}. In fact, when the data distribution is long-tailed, memorization might be necessary to achieve near-optimal accuracy on the test data \citep{feldman2020,brown2020memorization}. Leakage of memorized content can cause privacy violations, especially in the case where the content can be linked to an individual \citep{GDPR}.
\red{There is a wide range of data leakage detection and prevention techniques for document classification in the literature, e.g., \citep{alneyadi2013adaptable,katz2014coban,alneyadi2015detecting}. However, several challenges and limitations are identified with these techniques \citep{alneyadi2016survey,cheng2017enterprise}.}

In the case of language models trained with AR objective, the model learns to predict each and every next token given a sequence of tokens, which can theoretically lead to the leakage of the whole sequence if it is memorized by the model. \citep{carlini2021extracting} has shown a successful extraction of memorized data, including various personal information from the GPT-2 model \citep{RadfordWCLAS19} belonging to this family.

For language models trained with MLM objective, the story has been different so far. For instance, \citep{lehman2021does} shows that it is \emph{not} easy to extract sensitive information from the BERT model \citep{DevlinCLT19} trained on private clinical data. This can be attributed to the fact that the MLM objective only targets a small portion of \texttt{[MASK]} tokens randomly replaced in the training set, as opposed to all the tokens in the AR setting.

Other forms of privacy leakage include membership inference, which has been widely explored in vision and text scenarios \citep{shokri2017membership, yeom18, long18, truex18, song19, nasr19, sablayrolles19a, LOGAN, salem18, Leino20, choo2020labelonly, shejwalkar2021membership}, and property inference \citep{Ganju18, Wanrong21, chase2021property}. 

\section{This Work: Text Classification}
In this work, we turn our attention to the text classification setting, which spans a wide range of downstream applications \citep{Minaee21}. Often times pre-training a language model is performed on large public datasets while fine-tuning requires a much smaller task-specific dataset whose privacy requirements might be much more strict. To the best of our knowledge, this setting has been largely unexplored and our goal is to understand potential privacy leakage in this setting. 

In a text classification problem, the input is a sequence of tokens $\mathbf{x} = \left(x_1, x_2, \ldots, x_n\right)$ with a corresponding class label $y \in \left\{1,2, \ldots, C \right\}$ where $C$ is the number of classes. A model is trained to learn the relation between the input text and the corresponding class label. From a training data extraction perspective, the challenge of this setting is that here the goal is to maximize the log-likelihood of the correct class label (i.e. $\log \mathbb{P}\left(y | \mathbf{x}\right)$), therefore, there is no language modeling involved among the tokens of the sequence $\mathbf{x}$. Although we cannot leverage the approaches introduced in prior work, it is also not clear a priori whether one can extract training data given the partial knowledge of the tokens and the label with query access to the model.

\section{Threat Model and Testing Methodology}
Similar to prior work \citep{shokri2017membership, carlini2019secret}, we assume black-box access to the target model, where it receives a sequence of tokens and outputs a class prediction with its corresponding likelihood. 
Our goal is to investigate whether it is possible to extract the remaining tokens given partial information about a sequence under this black-box access to the target model.

This framework encompasses both a malicious attacker who has partial information about personal data points and aims to fully reconstruct it by fiddling with the target model, and any individual who audits a target model to detect any unauthorized use of personal data \citep{song19} (or to check whether a model owner has actually complied with data deletion requests). 
We choose to focus on the latter case since it allows the data owner to inject ``special'' sequences into their data that would strongly indicate unauthorized use of personal data if a successful reconstruction is possible through the target model.

Similar to \citep{thakkar2021}, we inject sequences of randomly selected tokens (with corresponding labels) into the training set. 
This mimics the existence of out-of-distribution data that is not pertinent to the learning process. We consider a testing procedure in which the goal of the {extraction algorithm} is to retrieve the last $n$ tokens of a canary\footnote{Since the model is bidirectional, this could be any arbitrary $n$ tokens in the sequence in general.}, where the sample space for each missing token is the entire tokenizer vocabulary. 
In the next section, we propose our extraction algorithm.

\section{Proposed Extraction Algorithm}
Given a partial sequence with missing tokens, the core idea of the proposed extraction algorithm is to choose the tokens such that the corresponding class label achieves the highest likelihood under the target model.
Consider a canary sequence $\mathbf{x} = \left(x_1, x_2, \ldots, x_n\right)$ with a corresponding label $y$. 
Given a partial input, we iteratively query the underlying classification model to reconstruct the missing tokens.
In particular, for a partial sequence $(x_1, x_2, \ldots, x_{t-1})$, the extraction algorithm enumerates all possible tokens from the vocabulary $\mathcal{V}$, evaluates the corresponding likelihood of the label $y$ for each token by querying the classification model, and then returns the token that achieves the maximum likelihood. Formally, $x_t$ is evaluated using the following optimization problem:
\begin{align}
    x_t = 
    \argmax_{v \in \mathcal{V}} \mathbb{P}\left(y | (x_1, x_2, \ldots, x_{t-1}, v) \right).
    \label{eq:xt_pred_1}
\end{align}

When a canary is repeated a few times in the training set, the extraction criterion in \eqref{eq:xt_pred_1} may not yield a successful reconstruction of the canary sequence. 
In order to boost the performance of token extraction, we propose a data-dependent regularizer to penalize the tokens with the highest number of occurrences in the training set, counteracting the model's bias towards these tokens.
Let $C(v)$ be the normalized number of occurrences of token $v$ in the training data\footnote{This may be a strong requirement but approximations can be made via publicly available datasets. However, the extraction performance does not degrade much by setting $\lambda=0$ (see Table~\ref{table:s1_lambda} in Section~\ref{sec:exp}).} for $v \in \mathcal{V}$.
Consequently, the optimization problem with the regularized objective function is given by 
\begin{align*}
    x_t \!=\! 
    \argmax_{v \in \mathcal{V}} \mathbb{P}\!\left(y | (x_1, x_2, \ldots, x_{t-1}, v)\!\right)
    \!-\! \lambda \!\cdot\! C(v),\!
\end{align*}
where $\lambda$ is the regularization coefficient that controls the amount of penalization imposed on the tokens with frequent occurrences in the training data.

\section{Experimental Evaluation}
\label{sec:exp}
\paragraph{Dataset:} We use the Reddit dataset\footnote{https://huggingface.co/datasets/reddit}.
We select the top 100~subreddits with largest number of reddit posts. We randomly sample 10000 and 2500 posts for the training and validation sets, respectively. 
The task is topic classification. In particular, given a user comment, the model is trained to predict the corresponding subreddit.

\paragraph{Model:} We use the pre-trained BERT base model \citep{devlin2019bert}. We fine-tune the model for 10 epochs using AdamW optimizer \citep{loshchilov2018decoupled} with weight decay 0.01, learning rate 1e-6, and batch size 32. We apply early stopping and take the snapshot that achieves the best validation performance to avoid overfitting.
\red{The average performance of the model over $10$ runs with different random seeds is as follows:}
\begin{itemize}[leftmargin=0.4cm]
    \setlength\itemsep{0em}
    \vspace{-1mm}
    \item The average training accuracy is $47.69\%$ for a training set size of $10$k samples.
    \item The average validation accuracy is $42.94\%$ for a validation set size of $2.5$k samples.
    \vspace{-2mm}
\end{itemize}

\begin{table}[t]
\centering
    \begin{tabular}{|c|c|c|c|c|}
        \hline
        \multicolumn{1}{|c|}{}  & \multicolumn{2}{|c|}{\small{Original Canary}} & \multicolumn{2}{c|}{\small{Supporting Canary}} \\
        \hline
        & \!\!\small{Subreddit}\!\! & \!\!\small{Repetitions}\!\! & \!\!\small{Subreddits}\!\! & \!\!\small{Repetitions}\!\!\\
        \hline
        \!\!\small{Table~\ref{table:s1_lambda}}\!\! & \!\!\small{Rarest}\!\! & \!\!\small{100}\!\! & \!\!\small{All Other}\!\! & \!\!\small{1}\!\!\\
        &  &  & \!\!\small{Subreddits}\!\! &\\
        \hline
        \!\!\small{Table~\ref{table:s1_bw_insertions}}\!\! & \!\!\small{Rarest}\!\! & \!\!\small{Varying}\!\! & \!\!\small{All Other}\!\! & \!\!\small{1}\!\!\\
        &  & \!\!\!\small{(1st Column)}\!\!\! & \!\!\small{Subreddits}\!\! &\\
        \hline
        \!\!\small{Table~\ref{table:s2_bw_insertions}}\!\! & \!\!\small{Rarest}\!\! & \!\!\small{100}\!\! & \!\!\small{One Other}\!\! & \!\!\small{Varying}\!\!\\
        &  &  & \!\!\small{Subreddit}\!\! & \!\!\!\small{(1st Column)}\!\!\! \\
        \hline
    \end{tabular}
    \caption{Information about the subreddits as well as numbers of repetitions of the original and supporting canaries for each experiment.} 
    \label{table:canaries}
    \vspace{-2mm}
\end{table}

\red{\paragraph{Canary Construction:} A canary sequence consists of a number of tokens and an associated class label. 
Each token in a canary is sampled uniformly at random from the BERT tokenizer vocabulary.
We exclude subwords and sample from the remaining $17$k whole words in the vocabulary.
The reason for random sampling of tokens is to construct out-of-distribution posts with very high probability.
For instance, an example of a randomly generated canary is ``expected Disney activated Fulton rebel scalp Stark fraud myths Palestine.'' 
Finally, a canary sequence is inserted into the training set and repeated multiple times.
This construction of canary sequences enables us to evaluate the model's unintended memorization of training data.}
\\

Intuitively, the most successful extraction is likely to occur within the rarest subreddit because there is more capacity for memorization. 
Hence, we insert a canary sequence of 10 randomly selected tokens into the rarest subreddit with 100 repetitions.
This will be the original canary for which we would like to perform token extractions.
Our first observation is that given the first 7 to 9 tokens, the model is already confident in the corresponding label, and hence the missing token(s) do not exhibit themselves in our optimization. 
In particular, $\mathbb{P}\left(y | (x_1, x_2, \ldots, x_{9}, v)\right)$ has similar values for all $v \in \mathcal{V}$.
Therefore, we inject one sequence into all other subreddits where the first 7 to 9 tokens are fixed, and the missing token(s) are chosen differently at random.
These are called \textit{supporting canaries} since they are not meant to be extracted, but enable the missing token(s) in the original canary to be crucial for maximizing the likelihood of the corresponding label, and hence the performance of the reconstruction is significantly boosted. 
\red{Table~\ref{table:canaries} shows detailed information about the original and supporting canaries for each experiment whose results are presented next}.
The success of reconstruction is defined by the appearance of the missing token(s) in the top-$k$ generation of the algorithm for a beam size $k$. 
Note that each experiment is run 10 times and the average success rate is reported. 
\red{In Table~\ref{table:s1_lambda}, we present the results of the aforementioned experiment with $k=100$}.
It is evident that the proposed algorithm achieves significant success rates for the extraction of a few tokens. 
However, it fails to reconstruct beyond more than two tokens since the search space becomes exponentially larger.

\begin{table}[t]
\centering
    \begin{tabular}{|c|c|c|c|}
        \hline
        \multicolumn{1}{|c|}{}  & \multicolumn{3}{c|}{\small{Success Rate}} \\
        \hline 
        $\lambda$ & \small{Last Token} & \small{Last 2 Tokens} & \small{Last 3 Tokens} \\
        \hline
        0 & {0.8} & {0.2} & 0 \\
        \hline
        \textbf{0.01} & \textbf{0.9} & \textbf{0.3} & 0 \\
        \hline
        0.1 & {0.7} & {0.1} & 0 \\
        \hline
        1 & {0.3} & {0.1} & 0\\
        \hline
        10 & {0.1} & 0 & 0 \\
        \hline
    \end{tabular}
    \caption{Successful extraction rates of the proposed algorithm on the last 1 to 3 tokens for different values of the regularization parameter $\lambda$. The original canary is inserted 100 times in the rarest subreddit, while the supporting canary is inserted only once in all other subreddits. Random guess rate is only $0.0058$ for the last token and $3.4$e-$5$ for last 2 tokens.} 
    \label{table:s1_lambda}
    \vspace{-2mm}
\end{table}

\begin{table}[t]
\centering
    \begin{tabular}{|c|c|c|c|}
        \hline
        \multicolumn{1}{|c|}{\!\!\small{Original Canary}\!\!} & \multicolumn{1}{|c|}{} & \multicolumn{2}{c|}{\small{Success Rate}} \\
        \hline 
        \small{Repetitions} & \!\!\small{Beam Size}\!\! & \!\!\small{Our Algo.}\!\! & \!\!\small{Random Guess}\!\! \\
        \hline
        100 & 50 & \textbf{0.7} & 0.0029\\
        \hline
        50 & 50 & \textbf{0.5} & 0.0029\\
        \hline
        25 & 50 & \textbf{0.1} & 0.0029\\
        \hline
        10 & 50 &  0 & \textbf{0.0029}\\
        \hline
        \hline
        100 & 100 & \textbf{0.9} & 0.0058\\
        \hline
        50 & 100 & \textbf{0.5} & 0.0058\\
        \hline
        25 & 100 & \textbf{0.3} & 0.0058\\
        \hline
        10 & 100 & \textbf{0.1} & 0.0058\\
        \hline
        \hline
        100 & 200  & \textbf{1} & 0.0117\\
        \hline
        50 & 200 & \textbf{0.9} & 0.0117\\
        \hline
        25 & 200 & \textbf{0.4} & 0.0117\\
        \hline
        10 & 200 & \textbf{0.2} & 0.0117\\
        \hline
    \end{tabular}
    \caption{Successful extraction rates of the proposed algorithm compared to random guessing on the last token for various repetitions of the original canary and beam sizes. The supporting canary is inserted only once in all subreddits except the rarest. We set $\lambda=0.01$.} 
    \label{table:s1_bw_insertions}
    \vspace{-2mm}
\end{table}

\red{Table~\ref{table:s1_bw_insertions} presents the extraction results for the last token for various repetitions of the original canary and beam sizes. The supporting canary is inserted only once in all subreddits except the rarest.}
Although high repetition improves the success rate of our algorithm, which aligns well with the findings that memorization is exacerbated by duplication of a sequence \citep{Kandpal22, carlini22}, low repetition still resurfaces the missing token if the algorithm generates a larger number of candidates (i.e., larger beam size). 

Instead of inserting one supporting canary into all subreddits except the rarest, we next investigate the insertion of a supporting canary into only one other arbitrarily chosen subreddit. 
\red{Here we fix 100 repetitions of the original canary in the rarest subreddit and vary the repetition of the supporting canary in a different subreddit}. 
Table~\ref{table:s2_bw_insertions} \red{shows the extraction results for this experiment for various repetitions of the supporting canary and beam sizes}.
We can see that extraction is possible even when a canary is inserted into the rarest subreddit only, as shown in the last part of Table~\ref{table:s2_bw_insertions}. However, the success rate improves greatly when we inject a supporting canary into another subreddit. The repetition we use for the subreddit does not seem to have an effect on the success rate of the extraction of the original canary.

\begin{table}[t]
\centering
    \begin{tabular}{|c|c|c|c|}
        \hline
        \multicolumn{1}{|c|}{\!\!\small{Supporting Canary}\!\!} & \multicolumn{1}{|c|}{} & \multicolumn{2}{c|}{\small{Success Rate}} \\
        \hline 
        \small{Repetitions} & \!\!\small{Beam Size}\!\! & \!\!\small{Our Algo.}\!\! & \!\!\small{Random Guess}\!\! \\
        \hline
        99 & 50 & \textbf{0.5} & 0.0029\\
        \hline
        99 & 100 & \textbf{0.5} & 0.0058\\
        \hline
        99 & 200 & \textbf{0.5} & 0.0117\\
        \hline
        \hline
        50 & 50 & \textbf{0.4} & 0.0029\\
        \hline
        50 & 100 & \textbf{0.4} & 0.0058\\
        \hline
        50 & 200 & \textbf{0.5} & 0.0117\\
        \hline
        \hline
        25 & 50 & \textbf{0.4} & 0.0029\\
        \hline
        25 & 100 & \textbf{0.5} & 0.0058\\
        \hline
        25 & 200 & \textbf{0.5} & 0.0117\\
        \hline
        \hline
        0 & 50 & \textbf{0.1} & 0.0029\\
        \hline
        0 & 100 & \textbf{0.1} & 0.0058\\
        \hline
        0 & 200 & \textbf{0.1} & 0.0117\\
        \hline
    \end{tabular}
    \caption{Success rates of extracting the last token under the proposed algorithm and random guess for various repetitions of the supporting canary and beam sizes.
    The original canary is inserted 100 times in the rarest subreddit. We set $\lambda = 0$.
    }
    \label{table:s2_bw_insertions}
    \vspace{-2mm}
\end{table}


\section{Conclusion and Future Work}
In this work, we studied the problem of unintentional memorization in a text classification setting. We developed an algorithm to extract unknown tokens of a partial text via access to the underlying classification model. Through experimental studies, we demonstrated the efficacy of the proposed extraction algorithm over random guessing.

Our experimental setting provides preliminary results and is subject to further exploration in future work. In particular, we injected the original canary into the rarest subreddit. In general, it would be interesting to range from the rarest to the most popular subreddit. We also used random tokens for canary construction, and it is of importance to extend it to more organic canaries. Finally, we leave investigating the effect of formal privacy guarantees, such as differentially private model training \citep{abadi2016deep}, to future work.

\section{Ethical Impact}
This work explores the privacy implications of a text classification setting in which training is performed on sensitive and private data. We investigate whether data leakage is feasible under this setting. 
We believe that this work is a first step in determining the susceptibility of the underlying text classification model to privacy leakage and detecting unauthorized use of personal data.
Both the dataset and the model are publicly available.

\section*{Acknowledgements}
We thank members of Microsoft's Privacy in AI (PAI) research group for the valuable feedback in the early stages of this work. Special thanks to Fatemehsadat Mireshghallah for the insightful discussions, and Mashaal Musleh for the support in setting up the experiment environment. Finally, we thank the anonymous reviewers for their helpful comments toward improving the paper.


\bibliography{myRef.bib}

\end{document}